\pdfoutput=1

\documentclass[11pt]{article}

\usepackage[preprint]{acl}

\usepackage{times}
\usepackage{latexsym}

\usepackage[T1]{fontenc}

\usepackage[utf8]{inputenc}

\usepackage{microtype}

\usepackage{inconsolata}

\usepackage{graphicx}

\usepackage[capitalise,noabbrev]{cleveref}
\usepackage{enumitem}
\usepackage{makecell}
\usepackage{subcaption}

\title{Efficient Annotator Reliability Assessment with EffiARA}

\author{
 \textbf{Owen Cook},
 \textbf{Jake Vasilakes},
 \textbf{Ian Roberts} and
 \textbf{Xingyi Song} \\
  School of Computer Science, \\ University of Sheffield \\
 \small{
   \textbf{Correspondence:} \{oscook1, x.song\}@sheffield.ac.uk
 }
}

\begin{document}
\maketitle
\begin{abstract}
Data annotation is an essential component of the machine learning pipeline; it is also a costly and time-consuming process. With the introduction of transformer-based models, annotation at the document level is increasingly popular; however, there is no standard framework for structuring such tasks. The EffiARA annotation framework is, to our knowledge, the first project to support the whole annotation pipeline, from understanding the resources required for an annotation task to compiling the annotated dataset and gaining insights into the reliability of individual annotators as well as the dataset as a whole. The framework's efficacy is supported by two previous studies: one improving classification performance through annotator-reliability-based soft-label aggregation and sample weighting, and the other increasing the overall agreement among annotators through removing identifying and replacing an unreliable annotator. This work introduces the EffiARA Python package and its accompanying webtool, which provides an accessible graphical user interface for the system. We open-source the EffiARA Python package at \url{https://github.com/MiniEggz/EffiARA} and the webtool is publicly accessible at \\ \url{https://effiara.gate.ac.uk}.
\end{abstract}

\section{Introduction}

Labelled data is the foundation of model training and evaluating downstream tasks in machine learning models. However, data annotation is often an expensive and time-consuming process, significantly affecting the quality of model training. Obtaining annotations from experts is ideal, but this expertise is often logistically and financially costly.

Crowd-sourcing platforms such as Amazon's Mechanical Turk\footnote{\url{https://www.mturk.com/}} and CrowdFlower (now Figure-Eight)\footnote{\url{https://www.appen.com/ai-data/data-annotation}} provide a cheaper alternative by using non-expert annotators; this generally results in lower quality annotations with higher levels of inter-annotator disagreement \citep{nowak2010reliable}. Effectively collecting, evaluating and managing annotator disagreement is essential in addressing challenges of data quality.

We introduce the EffiARA (Efficient Annotator Reliability Assessment) framework, which supports annotation quality assessment and management throughout the annotation process, allowing users to:

\begin{itemize}[topsep=2pt, itemsep=2pt, parsep=0pt]
    \item \textbf{Distribute} data points to annotators; 
    \item \textbf{Generate labels} from the annotations of each annotator;
    \item \textbf{Assess agreement} among annotators;
    \item \textbf{Assess annotator reliability};
    \item \textbf{Redistribute} data points to obtain the desired level of agreement;
    \item \textbf{Generate aggregated labels} at the data point level, taking either a soft- or hard-label approach. 
\end{itemize}

To our knowledge, no existing annotation framework provides systematic support for annotator workload allocation which can then be used to estimate the cost of the annotation project. This, in addition to the set of functionalities surrounding the annotation process, makes the EffiARA annotation framework a unique solution for structuring data annotation and modelling annotators.

Additionally, by aggregating annotators' labels for each data point, tempered by measures of annotator reliability, we can obtain a consensus that better reflects the ``true'' label distribution. Annotator reliability can also be used to dynamically weight individual data points during model training to ensure that the model prioritises reliable annotations \citep{cook2024efficient}.

\section{Related Work}

\subsection{Annotation Frameworks}
There have been many attempts to formalise the annotation process for a number of annotation tasks and a range of tools are available. Many frameworks focus on sequence-labelling tasks such as POS tagging and named-entity recognition \citep{bird2001formal, cornolti2013framework, bontcheva2013gate, lin2019alpacatag}. More recently, with the introduction of pre-trained LLMs capable of document-level processing, document annotation tools and frameworks have been created, such as GATE Teamware 2 \citep{wilby2023gate}. INCEpTION \citep{klie-etal-2018-inception, de2024integrating} handles various aspects of annotator management, including workload distribution, as well as the annotation itself, with a customisable UI, supporting span and document annotation. It also enables active learning but does not aim to model annotator or dataset reliability explicitly. A number of annotation frameworks are task-specific, aiming to provide a set of guidelines and tools for following them, for example event ordering \citep{cassidy2014annotation}, biodiversity information extraction \citep{lucking2022multiple}, and surgical video analysis \citep{meireles2021sages}.

\subsection{Annotator Agreement}

Agreement among annotators is often used to assess the quality of a dataset. Commonly used metrics include Scott's Pi \citep{scott1955reliability}, Cohen's Kappa \citep{cohen1960coefficient}, Fleiss' Kappa \citep{fleiss1971measuring}, and Krippendorff's alpha \citep{krippendorff1970estimating}. For each metric, there are various interpretations and accepted agreement thresholds used to determine the reliability of a dataset \citep{krippendorff2018content, landis1977measurement}. Obtaining datasets where this agreement threshold is met, particularly in scenarios with non-expert annotators such as crowd-sourcing, is challenging and costly \citep{hsueh2009data, nowak2010reliable}.

\subsection{Annotation Aggregation}
Rather than ensuring acceptable levels of agreement, many approaches use disagreement as additional information, utilising it to understand the subjectivity of particular data points or the reliability of annotators. 

The soft-label approach incorporates a level of subjectivity into aggregated labels for each data point and has been shown to improve both classification performance and model calibration \citep{wu2023don, cook2024efficient}. Popular methods of label aggregation include majority voting (hard-label only), \citet{dawid1979maximum}, GLAD \citep{whitehill2009whose}, and MACE \citep{hovy2013learning} for categorical data; these methods have been implemented in Python as part of the Crowd-Kit tool \citep{ustalov2021learning}.

\subsection{Annotator Reliability}
Assessing annotator reliability can be used to assess the quality of individual annotators and may be used to understand the quality of data available, remove low-reliability annotators \citep{cook2025dataset}, inform the training of machine learning models through aggregating soft labels with reliability \citep{dawid1979maximum, wu2023don}, or affect the loss function during training \citep{cao2023learning, cook2024efficient}. There are different approaches to assessing annotator reliability, such as learning through Expectation Maximisation \citep{cao2023learning} or directly inferring the reliability of an annotator from their agreement with others \citep{inel2014crowdtruth, dumitrache2018crowdtruth, cook2024efficient, cook2025dataset}. Through impacting the generation of soft labels, or directly impacting the training loss, more information about which annotations are more trustworthy is provided, leading to more performant and robust models. An alternative approach to assessing annotator reliability involves comparing annotators' annotations to a set of gold-standard labels \citep{barthet2023knowing}; this approach is often used to filter out bad annotators. All three approaches have been shown to improve model performance on classification tasks when compared to methods that trust each annotator equally.

\begin{figure*}[t]
    \centering
    \includegraphics[width=0.7\textwidth]{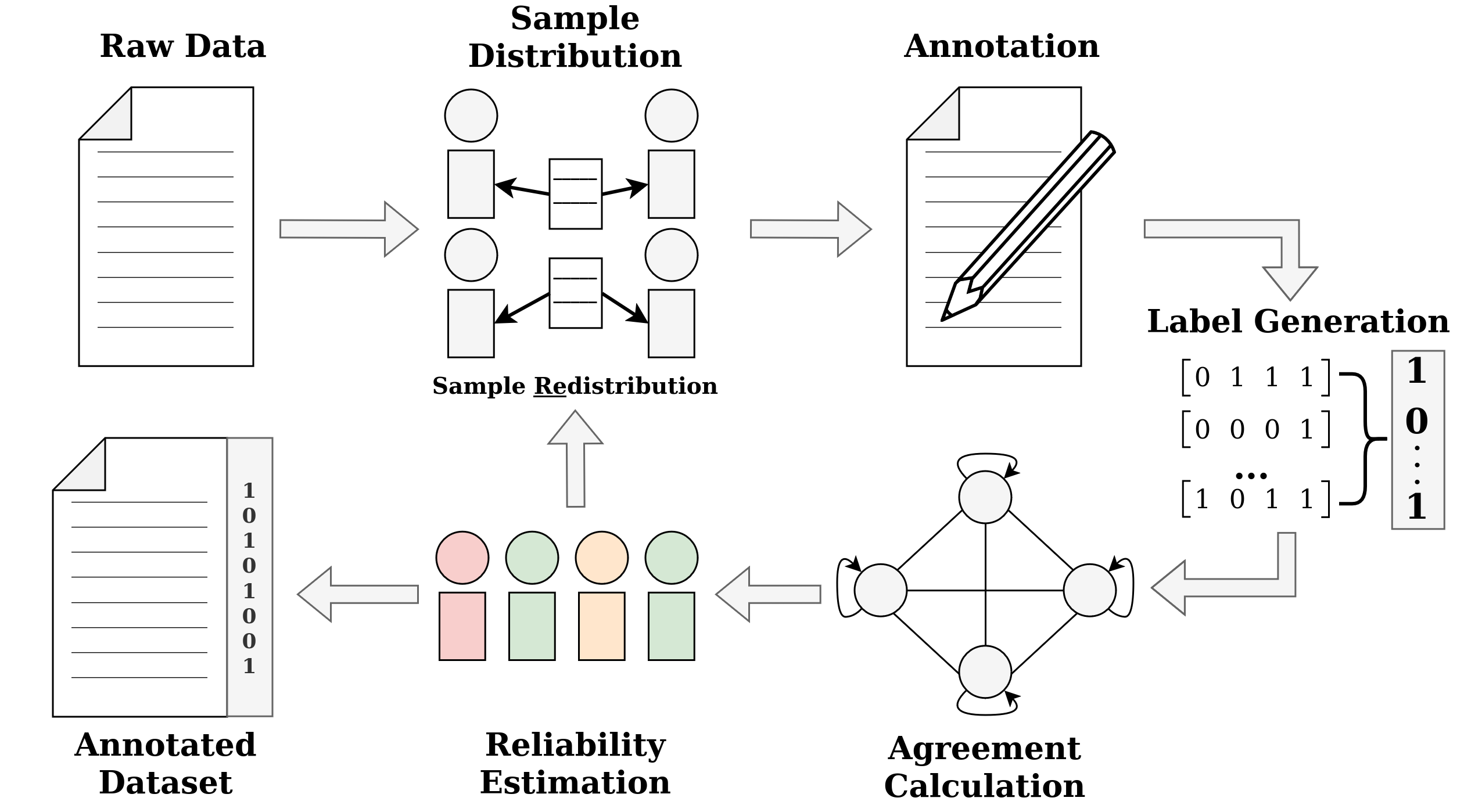}
    \caption{An overview of the EffiARA annotation pipeline, covering sample distribution, annotation, label generation, agreement calculation, reliability estimation, and dataset compilation.}
    \label{fig:pipeline}
\end{figure*}

\section{EffiARA Python Package}
The EffiARA annotation framework structures the annotation process from start to finish. It distributes samples to annotators, generates and aggregates labels, computes inter- and intra-annotator agreement, and assesses annotator reliability. A visual representation of the EffiARA pipeline is provided in \cref{fig:pipeline} and we describe each stage in detail below. The framework has been implemented in Python due to its extensive use in NLP research, increasing its ease of use and integration.

The annotation pipeline is implemented as a set of modular tools in the EffiARA Python package. The source code is available at \url{https://github.com/MiniEggz/EffiARA} and the package has been released on PyPi for quick installation: \url{https://pypi.org/project/effiara/}. Documentation is available here: \url{https://effiara.readthedocs.io}.

The package relies on a number of core Python libraries. Two fundamental libraries required by the EffiARA framework are NumPy \citep{oliphant2006guide, harris2020array} and pandas \citep{mckinney2011pandas}, used for efficient mathematical operations on arrays and the manipulation of data. 

\subsection{Sample Distribution}
The first stage in the EffiARA pipeline enables annotation coordinators to estimate resource requirements: how many annotators are needed, how much time is required from each annotator, and how many samples can be produced, given the time and number of annotators. Once resources have been finalised, data points can be distributed among annotators with the EffiARA distribution algorithm, which ensures annotator agreement can be effectively assessed \citep{cook2024efficient}. 

Both of these functionalities are implemented in the \texttt{SampleDistributor} class. We first use SymPy \citep{meurer2017sympy} to solve for the missing variable in the resource-understanding equation introduced in \citet{cook2024efficient} (Algorithm 1). We then use pandas to split the data into separate DataFrames for each annotator, with one DataFrame containing left-over samples that may be used later. 

\subsection{Data Annotation}
The sample allocations obtained in the previous step can then be used to assign samples to annotators and complete the annotation process using existing tools such as GATE Teamware 2 \citep{wilby2023gate} or Amazon's Mechanical Turk.

\subsection{Label Generation}
Label generation involves transforming raw annotations obtained from annotators into numeric encodings compatible with annotator agreement metrics (such as Cohen's Kappa, Fleiss' Kappa, Krippendorff's alpha, or cosine similarity) and model training. These transformations may be at the individual annotator level (for example, transforming first- and second-choice annotations into a categorical distribution), or at the data point level (aggregating annotations from multiple annotators).

As the exact transformations required are often task-specific, the abstract \texttt{LabelGenerator} class guides users to implement their own label generation code with three necessary methods: 
\begin{itemize}[topsep=2pt, itemsep=2pt, parsep=0pt]
    \item \texttt{add\_annotation\_prob\_labels} is used to represent each individual's raw annotations; 
    \item \texttt{add\_sample\_prob\_labels} is used to aggregate labels at the data point level, retaining disagreement in a soft label approach;
    \item \texttt{add\_sample\_hard\_labels} aggregates the annotations into a hard label, through methods such as majority voting or taking the maximum probability label from the aggregated soft label.
\end{itemize}
  For annotator agreement calculations, only \texttt{add\_annotation\_prob\_labels} must be implemented. To instantiate a class inheriting from \texttt{LabelGenerator}, the user must provide a list of annotator names and the label mapping: a dictionary where the key is the value represented in the DataFrame and the value is a numeric representation. This enables the extraction of individual annotations and their representation as a distribution across the available classes.

We provide a number of preset label generators: the \texttt{DefaultLabelGenerator}, for the cases in which no special label aggregation is necessary; the \texttt{EffiLabelGenerator}, mirroring the label generation and aggregation shown in \citet{cook2024efficient}; the \texttt{TopicLabelGenerator}, for multi-label tasks such as topic-extraction \citep{cook2025dataset}; and the \texttt{OrdinalLabelGenerator}, used for ordinal annotation tasks where a number of features are labelled on a scale. With the \texttt{label\_generator.from\_annotations} method, the specific class inheriting from \texttt{LabelGenerator} is instantiated from the raw annotations, requiring no additional coding from the user.

\subsection{Annotator Agreement \& Reliability}
\begin{figure}[ht]
\centering
\begin{minipage}[A]{0.48\linewidth}
    \centering
    \includegraphics[width=\linewidth]{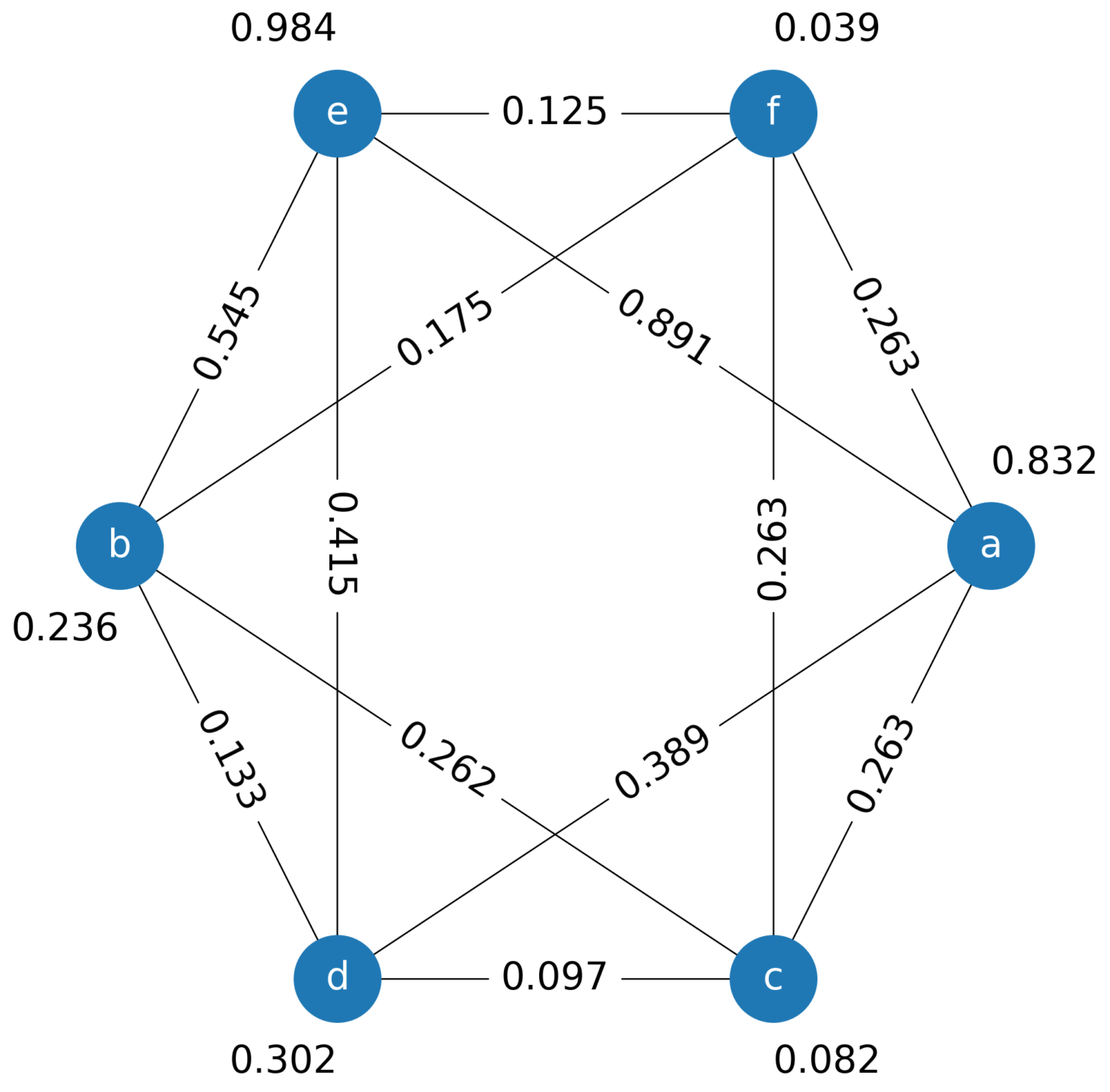}
    \subcaption{2D Graph}
\end{minipage}
\hfill
\begin{minipage}[B]{0.48\linewidth}
    \centering
    \includegraphics[width=\linewidth]{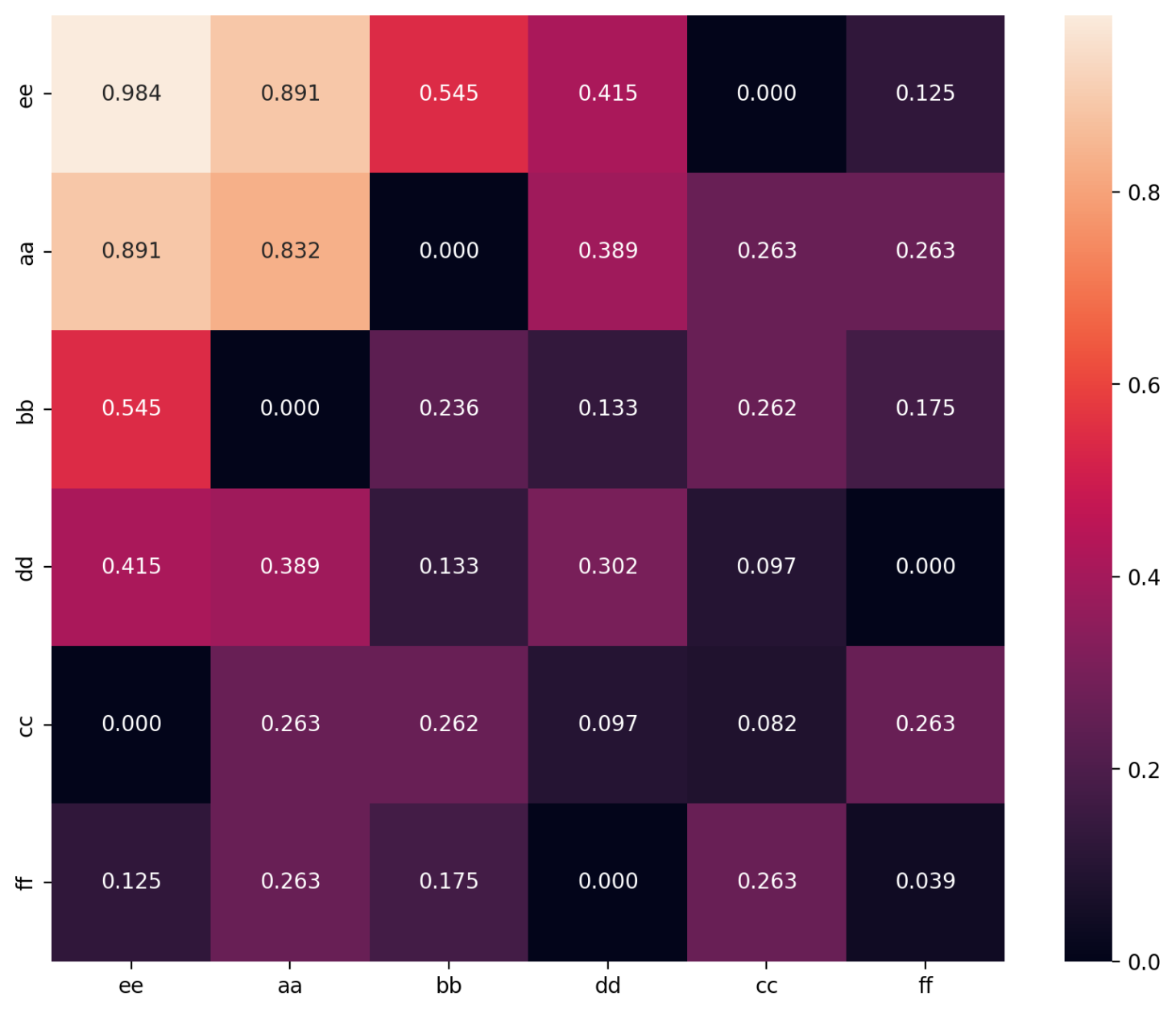}
    \subcaption{Heatmap}
\end{minipage}

\vspace{1em}

\begin{minipage}[C]{0.6\linewidth}
    \centering
    \includegraphics[width=\linewidth]{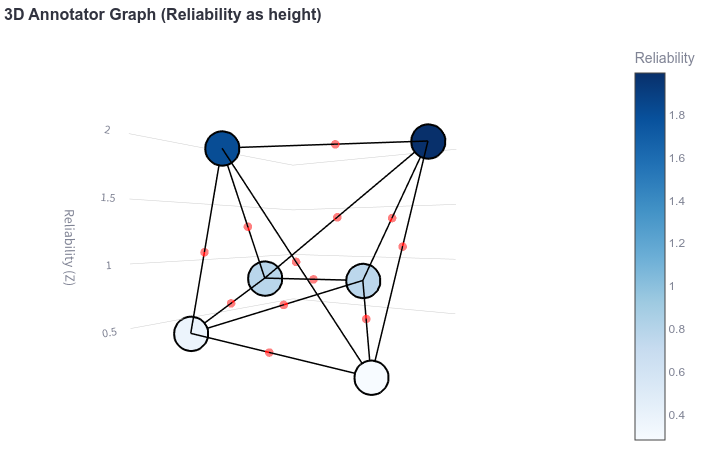}
    \subcaption{3D Graph}
\end{minipage}

\caption{Example agreement visualisations as (A) a 2D graph, (B) a heatmap, and (C) a 3D graph for six annotators (annotations were synthetically generated).}
\label{fig:agreement_visualisation}
\end{figure}

Once labels for each annotation have been generated, inter- and intra-annotator agreement are calculated using equations introduced in \citet{cook2024efficient}. Annotator agreement can then be visualised in a 2D or interactive 3D graph, where each node represents an annotator and edges between annotators represent the pairwise agreement between two annotators, with the value next to each node representing an annotator's agreement with themself. For cases with many annotators, where a graph could be unwieldy, we also provide a heatmap visualisation, where annotators are ordered by reliability; note that intra-annotator agreement is displayed on the diagonal. Examples of these visualisations are given in \cref{fig:agreement_visualisation}.

Using these agreement values, annotator reliability can then be calculated, using a combination of an annotator's intra-annotator agreement and average inter-annotator agreement, weighted by an $\alpha$ parameter controlling the strength of intra-annotator agreement from 0 to 1. The resulting agreement values are centered around 1, enabling the recursive inter-annotator agreement calculation from \citet{cook2024efficient}. The reliability values can then be accessed and utilised, potentially removing certain annotators from the annotation process \citep{cook2025dataset}. Reliability scores may also be utilised in label aggregation (in a \texttt{LabelGenerator}) or used to weight the loss function in model training \citep{cook2024efficient}.

Annotator agreement and reliability is calculated and stored in the \texttt{Annotations} class. The \texttt{Annotations} class is instantiated with a pandas DataFrame representation of the dataset, a \texttt{LabelGenerator} object (which will be generated using the \texttt{LabelGenerator.from\_annotations} function if no instance inheriting from \texttt{LabelGenerator} is passed), an agreement metric (defaulting to Krippendorff's alpha), an overlap threshold, and the reliability alpha. 

On instantiating an \texttt{Annotations} class, the annotator graph (supported by the NetworkX library \citep{hagberg2008exploring}) is initialised with each annotator equally reliable. Intra-annotator agreement is first calculated for each annotator node with the \texttt{calculate\_intra\_annotator\_agreement} instance method, using data points each user has annotated twice themselves. Inter-annotator agreement is then calculated between each user, utilising the \texttt{overlap\_threshold} to decide whether there is sufficient overlap between the two annotators to assess agreement. Here, the \texttt{pairwise\_agreement} function is used as a common interface to the implemented pairwise agreement metrics in the agreement module. Python modules used to handle agreement calculations include the Krippendorff library \citep{castro-2017-fast-krippendorff} for Krippendorff's alpha and Scikit-Learn \citep{pedregosa2011scikit} for Cohen's Kappa and Fleiss' Kappa. NumPy and pandas are also used for vector calculations and manipulation of the data to obtain pair annotations.

Once agreement has been calculated among annotators and with themselves, annotator reliability is calculated with a recursive application of the annotator reliability equation until reliability values converge. To ensure convergence, the calculated reliability values are normalised to have a mean of $1$ after each iteration. Annotator reliability values can then be accessed through the \texttt{get\_user\_reliability} and \texttt{get\_reliability\_dict} methods.

Inter- and intra-annotator agreement values can also be easily accessed via the graph itself using the NetworkX API and the \verb|__getitem__| method of the \texttt{Annotations} class. The graph and heatmap agreement visualisations shown in Figure \ref{fig:agreement_visualisation} utilise Matplotlib \citep{tosi2009matplotlib} and Seaborn \citep{waskom2021seaborn}, and they are displayed using the \texttt{display\_annotator\_graph} and \texttt{display\_agreement\_heatmap} methods respectively.

The optional \texttt{annotators} and \texttt{other\_annotators} arguments for the heatmap allow a user to display the agreement between one set of users and another, with the default setting comparing all annotators to one another. This may be useful in cases where you already have a set of reliable annotators or you have a gold-standard set of annotations you would like to compare a set of annotators to.

\subsection{Sample Redistribution}
In cases where a consensus must be reached on a high proportion of data points, samples may be redistributed among annotators to resolve disagreement. The \texttt{SampleRedistributor} provides this functionality. It functions very similarly to the \texttt{SampleDistributor} with the additional constraint that an annotator who has already annotated an individual data point will not be reassigned it. Sample redistribution can be done iteratively until the desired level of agreement is reached.

The \texttt{SampleRedistributor} inherits from the \texttt{SampleDistributor} class, overloading the \texttt{distribute\_samples} method, applying a round-robin-style allocation using the EffiARA sample distribution variables, ensuring that annotators are not given samples they have already annotated.

\subsection{Final Dataset}
Once the desired level of agreement has been reached, potentially with the aim of generating gold-standard labels in classification tasks, the final dataset is ready, with annotations tied to annotator identities, allowing for training strategies that utilise the expertise and reliability of individual annotators. Users may utilise the \texttt{concat\_annotations} method in the \texttt{data\_generation} module for assistance in merging annotations into the final dataset.

\section{EffiARA Webtool}
To make the functionalities of the EffiARA package more accessible and quicker to use, we have also released the webtool at \url{https://effiara.gate.ac.uk}. The webtool allows non-technical experts to run annotation projects and gain insights into annotator agreement and reliability with ease. Even for those comfortable using the Python package, the webtool provides a convenient interface for performing tasks quickly. A system demonstration is available at \url{https://www.youtube.com/watch?v=KcmQfPiskcY}.

The webtool supports common tasks within the annotation pipeline (excluding the annotation step itself). Finer-grained control and more advanced functionality can be achieved with the Python package, particularly through customisation of modules like the \texttt{LabelGenerator}. As the project is open-sourced, technical users are able to make their own modifications and run them as a local web-application or make a pull request to add their additional use-cases. The webtool source code is available at \url{https://github.com/MiniEggz/EffiARA-webtool}.

The application contains four main workflows: 

\begin{itemize}[topsep=2pt, itemsep=2pt, parsep=0pt]
    \item \textit{Sample Distribution.} This workflow handles all aspects of distributing samples from an unannotated dataset, including understanding the resources available. The \texttt{sample\_id} column is added to each data point to allow recompilation after annotation.
    \item \textit{Annotation Project.} This workflow is used to generate an annotation project for specific platforms. Currently, project generation for GATE Teamware 2 \citep{wilby2023gate} is supported. Future iterations may include other platforms but this task is most likely solved to some extent by the individual annotation platforms. 
    \item \textit{Dataset Compilation.} Once data annotation is complete, this  workflow allows the user to upload a ZIP file containing all annotation CSV files. It supports users in renaming columns, moving all reannotations under the correct columns (beginning with \texttt{re\_}) and into the correct row (alongside their original annotation of the data point), and merging the annotations from different annotators to create a final dataset ready for analysis.
    \item \textit{Annotator Reliability.} With the compiled dataset, users can analyse annotator reliability. The user first selects their label generator and they then have full control over the label mapping or they may choose to generate it automatically using the \texttt{LabelGenerator.from\_annotations} method. Users then choose the desired output: any combination of outputting annotator reliability, the annotator agreement graph (in 2D or interactive 3D) and an annotator heatmap. The workflow also offers a number of options for calculating annotator reliability, such as the agreement metric, the reliability alpha, and the overlap threshold (the minimum number of data points annotated by both annotators to enable agreement assessment); the workflow also offers display configurations for the graphs.
\end{itemize}

The webtool is built upon the EffiARA Python package and shares the same dependencies. It is implemented using Streamlit \citep{khorasani2022web} and Plotly \citep{sievert2020interactive} is used to create the interactive 3D annotator agreement and reliability visualisation. The \texttt{zipfile} and \texttt{tempfile} libraries handle uploads and downloads, ensuring data is deleted once processed.

\section{Evaluation}

\subsection{Case Studies}
Two previous works involving dataset creation have annotated data following the EffiARA methodology, creating RUC-MCD \citep{cook2024efficient} and the Chinese News Framing dataset \citep{cook2025dataset}. Both studies provide support for the annotation framework.

\paragraph{RUC-MCD.} In the work introducing the EffiARA annotation framework \citep{cook2024efficient}, utilising reliability scores in the label generation and model training stages was shown to improve classification performance. Applying a soft-label approach, using TwHIN-BERT-Large, assessing reliability with inter-annotator agreement only, intra-annotator agreement only, and a combination of both all improved classification performance. Classification performance increased from an F1-macro score of 0.691 to 0.740 using the EffiARA reliability scores calculated using a reliability alpha of $0.5$. The dataset used in this study was of low-to-moderate agreement, highlighting the framework's utility in datasets containing disagreement.

\paragraph{Chinese News Framing.} This work utilises the EffiARA reliability scores to identify unreliable annotators during the annotation process, leading to an increased overall level of agreement among annotators, which is highly indicative of data quality \citep{krippendorff2018content}. By removing the low-reliability annotator and replacing them with an existing high-reliability annotator, the average inter-annotator agreement (measured using Krippendorff's alpha) increased from $0.396$ to $0.465$. 

\subsection{Load Testing}
To assess the usability of the application, we also carried out load testing on the web application when hosted locally on a laptop with an Intel i7-6600U @ 3.400GHz and 16GB RAM, meaning upload and download speed were not a factor. Sample distribution remains quick and responsive for a large number of samples, taking less than a second for datasets of 100,000 samples. Dataset compilation and processing both scale roughly linearly with respect to dataset size with the tool requiring significantly longer to process datasets containing as many as 100,000 data points. Datasets containing 10,000 data points and under require less than one minute for dataset compilation and dataset processing (including annotator reliability calculation and visualisation rendering). The time taken for each key action in the webtool can be seen in \cref{tab:effiara_performance}. While running tasks that take longer, the web application remains responsive.

\begin{table}[ht]
\centering
\small
\begin{tabular}{|c|c|c|c|}
\hline
\makecell[c]{\textbf{Number of} \\ \textbf{Samples}} & \makecell[c]{\textbf{Sample} \\ \textbf{Distribution}} & \makecell[c]{\textbf{Dataset} \\ \textbf{Compilation}} & \makecell[c]{\textbf{Dataset} \\ \textbf{Processing}} \\
\hline
500   & \textasciitilde0.06s & \textasciitilde3s & \textasciitilde3s \\
\hline
1,000   & \textasciitilde0.06s & \textasciitilde6s & \textasciitilde6s \\
\hline
5,000   & \textasciitilde0.10s & \textasciitilde30s & \textasciitilde25s \\
\hline
10,000  & \textasciitilde0.12s & \textasciitilde1m & \textasciitilde45s \\
\hline
100,000  & \textasciitilde0.5s & \textasciitilde10m & \textasciitilde7m 20s\\
\hline
\end{tabular}
\caption{Processing time for each stage at varying dataset sizes. Tests conducted running the webtool locally on a laptop with 16GB RAM and an i7-6600U @ 3.400GHz.}
\label{tab:effiara_performance}
\end{table}

\section{Conclusion and Future Work}
In this work, we introduced the EffiARA Python package alongside an accessible web application that provides a graphical interface to the EffiARA annotation framework. EffiARA supports the design, compilation, and reliability assessment of annotation projects at the document level. 

Future development will focus on expanding the range of supported annotation settings, optimising computational performance, and enhancing usability based on user feedback. The package and webtool will be actively maintained, ensuring they remain usable and up-to-date with users' annotation requirements.

\section{Limitations}
EffiARA has been developed primarily to support document classification annotation, with the flexibility to accommodate other annotation types. While the design accounts for such flexibility, span annotation functionality, for example, has not yet been implemented and would require technical expertise to integrate. One key future addition may include the integration of Krippendorff's unitising alpha \citep{krippendorff2016reliability} for spans.

Some annotation tasks may require task-specific label generators. Although the framework includes preset label generators, less common tasks may require further customisation. This may present a usability barrier for non-technical users. We aim to address this through future development and community contributions of further task-specific components.

While EffiARA has been tested in situations with disagreement, it has been assumed that there is a ground-truth label for each data point. Future work will investigate the extension of EffiARA to tasks where there may not be a single ground-truth label, but potentially multiple subjective, and equally valid, true labels.

\section{Ethical Impact}
As EffiARA is an annotation framework, it does not pose direct ethical risks. Annotated data is instrumental in training machine learning models, including those that may be deployed in sensitive or high-impact contexts. Users of the EffiARA annotation framework should remain aware of the broader ethical impact of their annotation projects and consider them before undertaking such projects.

\section{Acknowledgements}
This work was supported by the UK’s innovation agency (Innovate UK) grant 10039055 (approved under the Horizon Europe Programme as vera.ai, EU grant agreement 101070093) vera.ai. JV is supported by the European Media and Information Fund (EMIF) managed by the Calouste Gulbenkian Foundation\footnote{The sole responsibility for any content supported by the European Media and Information Fund lies with the author(s) and it may not necessarily reflect the positions of the EMIF and the Fund Partners, the Calouste Gulbenkian Foundation and the European University Institute.} under the ``Supporting Research into Media, Disinformation and Information Literacy Across Europe'' call (ExU -- project number: 291191).\footnote{\url{exuproject.sites.sheffield.ac.uk}} We also acknowledge the IT Services at The University of Sheffield, utilising their High Performance Computing clusters in both of our case studies.

\bibliography{custom}

\end{document}